\pdfoutput=1

\documentclass[11pt]{article}

\usepackage[]{acl}

\usepackage{times}
\usepackage{latexsym}

\usepackage{hyperref}
\usepackage{url}
\usepackage{multirow}
\usepackage{booktabs}
\usepackage{enumitem}
\usepackage{xcolor}
\usepackage{amsmath}
\usepackage{flexisym}
\usepackage{bm}
\usepackage{tikz}
\usepackage{subfig}

\usepackage{float}
\usepackage{breqn}
\usepackage{dialogue}
\usepackage{listings}

\usepackage[T1]{fontenc}

\usepackage[utf8]{inputenc}

\usepackage{microtype}

\usepackage{inconsolata}
\newcommand{\kkcomment}[1]{}

\title{A Study on Leveraging Search and Self-Feedback for Agent Reasoning}

\author{
    Karthikeyan K$^{1}$\thanks{Work done as an intern at AWS},
    Michelle Yuan$^{2}$,
    Elman Mansimov$^{2}$,
    Katerina Margatina$^{2}$ \\
    \textbf{Anurag Pratik}$^{2}$,
    \textbf{Daniele Bonadiman}$^{2}$,
    \textbf{Monica Sunkara}$^{2}$,
    \textbf{Yi Zhang}$^{2}$,
    \textbf{Yassine Benajiba}$^{2}$
    \\
    $^{1}$Department of Computer Science, Duke University\\
    $^{2}$Amazon Web Services\\
    \texttt{karthikeyan.k@duke.edu}\\
    \texttt{\{miyuan,mansimov,katemarg,anuragik,dbonadim,sunkaral,yizhngn\}@amazon.com}
}

\begin{document}

\maketitle

\begin{abstract}
Recent works have demonstrated that incorporating search during inference can significantly improve reasoning capabilities of language agents.  Some approaches may make use of the ground truth or rely on model's own generated feedback. The search algorithm uses this feedback to then produce values that will update its criterion for exploring and exploiting various reasoning paths. In this study, we investigate how search and model's self-feedback can be leveraged for reasoning tasks. First, we explore differences in ground-truth feedback and self-feedback during search for math reasoning. Second, we observe limitations in applying search techniques to more complex tasks like tool-calling and design domain-specific approaches to address these gaps. Our experiments reveal challenges related to generalization when solely relying on self-feedback during search. For search to work effectively, either access to the ground-truth is needed or feedback mechanisms need to be carefully designed for the specific task. 
\end{abstract}

\section{Introduction}

Search algorithms have traditionally relied on ground-truth feedback, particularly in domains like games where win/loss conditions provide clear signals for decision-making~\cite{knuth1998}. Some recent works have shown potential in incorporating search during model inference to improve reasoning~\cite{hao-etal-2023-reasoning, zhou2023language,zhang2024accessinggpt4levelmathematical}. Typically, the search process requires feedback on the correctness of the candidate solutions. While such feedback traditionally came from external verification, these recent works have explored using the model's own assessment as feedback during search. The use of self-feedback is motivated by emerging evidence of agent's capability for self-correction and self-refinement~\cite{weng2023large,chen2025sets}. This direction is particularly exciting as it suggests the potential for models to guide their own search process and evolve without relying on external verification, moving us closer to more generalized, autonomous agents~\cite{putta2024agent}.

However, when implementing search for reasoning, a critical question emerges: how valuable is the model's own feedback in guiding the search process? While models may be capable judges in some scenarios~\cite{zheng2023judging}, their ability to reliably assess their own outputs during search remains questionable. This becomes particularly important when ground-truth feedback is unavailable, as is often the case in real-world applications.

In this study, we investigate this question through two distinct tasks: mathematical reasoning, GSM8K~\cite{cobbe2021gsm8k}, and tool-calling, ToolTalk~\cite{farn2023tooltalk}. Our analysis on GSM8K reveals that while search itself is valuable, using the model's self-feedback to guide the search may not be optimal. Instead, other strategies, like majority voting across multiple nodes during search, proves to be more effective.
We then extend our investigation to tool-calling, a more complex domain involving strategic decision-making and parameter selection. Here, we find that search with self-feedback can actually degrade performance. This leads us to explore domain-specific approaches for augmenting feedback mechanisms, including in-content examples and specialized verification modules.

While the promise of self-improving models through feedback and search remains appealing, the current reality of using self-feedback in search needs to be carefully reconsidered. This highlights a gap between models' previously demonstrated self-correction capabilities in some scenarios and their reliability in guiding search processes for reasoning. Overall, this suggests the need for more engineered feedback mechanisms tailored to specific tasks~\cite{zheng2025monte} or alternative approaches for leveraging self-feedback outside of search~\cite{chen2025sets}. These findings are aligned with recent generative AI breakthroughs that also mention the limitations of search and self-feedback for agent reasoning~\cite{guo2025deepseek}.

\section{Related Work}

Recent research has extensively explored methods to enhance LLM performance through advanced test-time compute approaches. \citet{wu2024inference} investigates scaling behaviors of sampling strategies, while \citet{muennighoff2025s1} show positive results with a simple test-time scaling approach called budget forcing.
Recent work highlight emerging potential for self-feedback~\cite{weng2023large,chen2025sets}. Likewise, there is also critique on whether models can truly evaluate and correct their own outputs~\cite{stechly2024selfverificationlimitationslargelanguage,kambhampati2024position}. 

Beyond sampling and iterative refinement, other works propose using Monte Carlo Tree Search (MCTS) as a more structured approach to improve reasoning~\cite{hao-etal-2023-reasoning, zhou2023language,zhang2024accessinggpt4levelmathematical}. MCTS can explore the search space effectively, trading off exploration and exploitation using the $UCT$ criterion: 
\begin{equation}
    UCT(a) = Q(s, a) + w\sqrt{ \frac{\log{N(s)} }{\log N(c(s, a)) } } \label{eq:UCT}
\end{equation}
where $Q(s, a)$ is the Q-value of taking an action $a$ from node $s$. $N(s)$  and $N(c(s, a))$ is the number of visits to node $s$ and its children, $c(s, a)$ respectively.  In  expansion and simulation (rollouts) stages, a new node is created and the Q-value, corresponding to the action that created the node, is initialized with rewards and updated during backpropagation. These rewards will be based on some source of feedback, whether it is from the ground-truth or another source of verification. Thus, the quality of this feedback is crucial for search to operate successfully.

\begin{table*}[t] 
    \centering 
    \begin{tabular}{lrrrrr}
        \toprule 
        & Llama 3 & Mistral v0.3 & Haiku 3  & Sonnet 3 \\
        \midrule 
        No Search & 0.813 & 0.426 & 0.866  & 0.757 \\
        MCTS: Ground-truth Feedback &  0.958  &  0.82  & 0.964  &  0.923  \\
        \midrule
        MCTS: Random Selection & 0.751 & 0.45  & 0.864 &  0.680 \\
        MCTS: Majority Voting & \textbf{0.883}  & \textbf{0.608} & \textbf{0.905}  & \textbf{0.786} \\
        MCTS: Maximum Reward & 0.776 & 0.469 & 0.854 & 0.685 \\
        \bottomrule 
    \end{tabular}
    \caption{Experiment results on math reasoning dataset GSM8K. The first two rows of the table correspond to the baselines: 1) no search, 2) original MCTSr implementation with ground-truth feedback. The last three rows correspond to our proposed modifications to test the effectiveness of self-feedback in search. Across all models, there is at least $\approx$ 10\% improvement in accuracy using the original MCTSr implementation with ground-truth verification over no search. However, without access to ground-truth feedback, the most promising alternative seems to be majority voting rather than picking the answer with maximum reward given from self-feedback.}
    \label{table:unrealistic}
\end{table*}

\section{Search without Ground-truth Feedback}~\label{sec:unrealistic_assumptions}
In this section, we explore the use of the model's own feedback for search on math reasoning dataset GSM8K. We build on the MCTSr (MCTS with Self-Refine) framework~\cite{zhang2024accessinggpt4levelmathematical}, which computes $Q$ value based on the model's feedback of its generated solution. One caveat with MCTSr is that the method originally assumes access to ground-truth feedback for early stopping and answer selection. Access to ground-truth feedback is appropriate for games like chess and Go. In other scenarios, there may be a lack of access to ground-truth feedback. This raises the question: Can we still leverage the model's self-feedback or other signals to guide search effectively? To address this, we investigate alternative strategies for answer selection and evaluate their performance in the absence of ground-truth feedback.

\subsection{Experimental Setup}
\label{sec:gt_exp}

We evaluate our approaches on the GSM8K dataset, which has become a standard benchmark for assessing mathematical reasoning capabilities in language models. Our experiments involve both closed-source and open-source language models, including Llama 3 Instruct (70B), Mistral v0.3 (7B), Claude 3 Haiku, and Claude 3 Sonnet.

For each model, we first establish a baseline performance through direct generation without any search mechanisms. We then run MCTSr with ground-truth verification to establish an upper-bound on performance, representing the best-case scenario where ground-truth feedback is available. Finally, we run MCTS using our proposed selection strategies that do not require ground-truth feedback: 
\textbf{1) Random selection:} A node from the search tree is chosen at random as the final answer. This approach serves as a baseline to measure the effectiveness of more sophisticated selection methods. 
\textbf{2) Majority voting:} We group the final numerical answers from all nodes in the search tree and select the answer that appears most frequently. This strategy aggregates the model's predictions, assuming that the most common answer is likely to be correct.
\textbf{3) Maximum reward:} The node with the highest self-feedback reward score is selected as the final answer. This approach relies on the model's ability to evaluate its own solutions, assuming that higher reward scores correspond to better answers. 

For the proposed selection strategies, we do not perform any early stopping and instead conduct search for a maximum of 10 MCTS iterations.
In Table~\ref{table:unrealistic}, we report the accuracy on GSM8K with no-search baseline, MCTSr with access to ground-truth feedback, and the proposed alternatives discussed above. In Appendix~\ref{sec:MCTSr_experiments}, we report detailed results along with other aggregation strategies like based on average rewards or weighted majority voting.

\subsection{Analysis} 

In Table~\ref{table:unrealistic}, ground-truth verification plays a huge role in the observed performance improvement. Across all models, there is at least $\approx $ 10\% improvement in accuracy using the original MCTSr implementation with ground-truth verification over no search. Within the strategies that do not rely on ground-truth feedback, majority voting seems to be the only selection strategy that consistently improves over the no-search baseline. Reward-based and random selection strategies seems to slightly improve the performance for some models and worsen for others. This indicates that self-feedback may not be a reliable source for providing rewards to select answers during search.

\section{Search with Augmented Feedback}
This section explores the application of search to a more complex domain of tool-calling, specifically using the ToolTalk dataset~\cite{farn2023tooltalk}. Unlike math reasoning tasks where there is one answer and verification is relatively straightforward, ToolTalk presents a significantly more nuanced challenge. The dataset consists of multi-turn dialogues where agents must understand user intents, decide when to make tool calls versus asking for clarification, and ensure all tool parameters are grounded in the conversation context. ToolTalk evaluation is done using teacher forcing, where we condition the conversation history based on the ground-truth conversation and evaluate the agent's response.
Within each turn, there can be zero to multiple tool calls, and ToolTalk evaluation sequentially decodes one tool call at a time. If the agent response contains text without tool calls, it's considered turn completion; if it contains a tool call, the tool is executed with the provided parameters, and the result is given back to the agent for continued generation.

\begin{table}[t] 
    \centering 
    \begin{tabular}{p{2cm}p{1.1cm}p{1.1cm}p{1.1cm}} 
        \toprule 
         & Precision & Recall & F1  \\
        \midrule 
        \multicolumn{4}{c}{Sonnet 3} \\
        \hline
        No Search   & 0.656  &  0.765& 0.706  \\
        MCTS &  0.502 &  0.630 & 0.559 \\
        \midrule
        \multicolumn{4}{c}{Haiku 3} \\
        \hline
        No Search   & 0.588 & 0.698 & 0.638  \\
        MCTS &  0.567 & 0.648 & 0.605 \\
        \bottomrule 
    \end{tabular}
    \caption{We compare using and not using search for tool-calling dataset ToolTalk. We report precision, recall, and F1 scores, which are averaged across 5 runs. For search, we use MCTS with maximum reward aggregation strategy based on self-feedback as described in Section~\ref{sec:gt_exp}. MCTS based on self-feedback seems to worsen the performance on ToolTalk.} 
    \label{tab:table_mctsr_tooltalk}
\end{table}

\begin{table*}[t] 
    \centering 
    \begin{tabular}{p{1.5cm}*{9}{p{0.05\textwidth} }} 
        \toprule 
         & \multicolumn{3}{c}{Augmentation with Guidelines} & \multicolumn{3}{c}{Augmentation with ICL} & \multicolumn{3}{c}{Augmentation with Module} \\
         \midrule
         & P & R & F1 & P & R & F1 & P & R & F1 \\
        \midrule 
        Sonnet 3 &  0.532 & 0.606 & 0.566 & 0.708 & 0.671 & \textbf{0.689} & 0.754 & 0.544 &  0.632 \\
        Haiku 3 & 0.547 & 0.622& 0.582 & 0.622 & 0.608 & 0.615 & 0.623 & 0.709 & \textbf{0.663}\\
        \bottomrule 
    \end{tabular}
    \caption{On ToolTalk, we compare three strategies to build better feedback mechanisms, namely (1) augmenting system prompt with guidelines on detecting hallucinations, (2) augmenting prompt with ICL examples, (3) augmenting feedback with an additional module to detect hallucinations. Again, we report precision, recall, and F1 scores averaged across 5 runs.} 
    \label{tab:tooltalk_improved}
\end{table*}

The complexity of ToolTalk stems from its open-ended nature and the strategic decisions required at each turn. Agents must not only understand what tools are available but also determine the appropriate moment to use them. A successful response often requires maintaining coherence across multiple turns while avoiding a common pitfall: parameter hallucination, where models fabricate plausible but incorrect tool parameters. This represents a fundamental departure from math reasoning tasks, where the challenge lies primarily in computational logic rather than strategic decision-making.

\subsection{Gaps in Search for Tool Calling}

We follow the same evaluation setup as ToolTalk. We compare a no-search baseline against using MCTS with self-refine. In this setup, each node represents a complete solution generated by the agent with tool-calling functionality, accompanied by its own feedback using a generic system prompt for obtaining feedback. Note that we do not execute tools during the search process, as some tool executions in real-world scenarios can have irreversible impacts. After completing the search, we iterate through all nodes in the tree and select the one with the highest reward.

\textbf{Analysis:} In Table~\ref{tab:table_mctsr_tooltalk}, we compare no-search baseline with self-refine search. We observe that the no-search baseline performs better than search, which indicate that self-refine could be detrimental~\cite{huang2024largelanguagemodelsselfcorrect}. We observe tool parameter hallucination as a major cause of error (please refer to Appendix~\ref{sec:appendix_tooltalk_errornalysis} for examples). In particular, when the user asks the agent to perform a task with insufficient information, instead of following up, the agent exhibits a bias towards making premature tool call requests with incomplete or hallucinated tool parameters~\cite{shaikh-etal-2024-grounding}. 
For example, when the user asks to register an account, the agent often hallucinates dummy credentials such as  "newuser" or "newpassword". Moreover, even the feedback and reward model does not capture these hallucinations in tool parameters.

\subsection{Augmenting Feedback with More Sources}

To mitigate poor feedback quality in search, we explore three strategies:
\textbf{1) Augmentation with Guidelines:} We refine the feedback model's system prompt and instruct it to specifically penalize hallucinations. 
\textbf{2) Augmentation with ICL examples:} We manually annotate a few in-context examples for the feedback model where these examples illustrate both hallucinated and factual agent responses, along with their appropriate rewards. 
\textbf{3) Augmentation with Hallucination Detection Module:} We augment the feedback with a separate hallucination detection module.  This module iterates through each tool parameter and asks the model if the parameter is provided by the user. If the answer is no, then it is considered as hallucinated parameter. After we iterate through all the parameters individually, we aggregate them to form the hallucination decision. Finally, we pass the agent response along with this hallucination decision to the feedback model to generate an overall feedback. We report precision, recall and F1 scores for all three strategies (Table~\ref{tab:tooltalk_improved}). Please refer to Appendix ~\ref{sec:appendix_tooltalk_systemprompts} for system prompts.

\textbf{Analysis:} Table~\ref{tab:tooltalk_improved} shows refining the system prompt is not effective as the results deteriorated compared to generic system prompt. Including in-context examples helps, especially with precision. Finally, with the hallucination detection module, precision increases significantly but recall drops significantly as well. When inspected carefully, we observe less hallucinations but the selected responses are prone to ask user for confirmation or unnecessary information. For example, when a user asks to delete an account, even if all the required information is present the model asks the user to confirm. Please refer to Appendix~\ref{sec:appendix_tooltalk_errornalysis} for a few examples illustration this behaviour. 


\section{Conclusion}

In this work, we conduct a study on integrating self-feedback into search for agent reasoning. While search remains a valuable technique for enhancing model performance, our results demonstrate that relying on self-feedback may be suboptimal or even detrimental in certain contexts. 
These insights have implications for the development of generalized, autonomous agents. Rather than pursuing purely self-guided approaches, our work indicates that successful search implementations may require carefully designed, domain-specific feedback mechanisms or hybrid approaches that combine self-refinement with other verification strategies.

\section{Limitations}

As a short paper, we have limited our scope to specific search methods (MCTS) and domains (math reasoning and tool calling). This focused approach allows for in-depth analysis within our chosen contexts. Future research could build on these findings by exploring additional reasoning domains, search algorithms, and datasets, potentially uncovering more patterns in self-feedback across various reasoning tasks. More exploration can also be done on augmenting self-feedback for search through other approaches.

\bibliography{anthology,custom}

\appendix


\section{MCTSr Experiments on GSM8K} \label{sec:MCTSr_experiments}

In Table~\ref{tab:mctsr_appendix}, we report results on GSM8k dataset with MCTSr search method with various strategies in place of ground truth feedback. 

\begin{table*}[t] 
    \centering 
    \begin{tabular}{p{5.5cm}p{2.1cm}p{2.1cm}p{2.1cm}p{2.1cm}} 
        \toprule 
        & LLaMA 3 & Mistral v0.3 & haiku 3  & Sonnet 3 \\
        \midrule 
        MCTS: Ground-truth Feedback &  0.958  &  0.82  & 0.964  &  0.923  \\
        No Search & 0.813 & 0.426 & 0.866  & 0.757 \\
        \hline
        MCTS: Random Selection & 0.751$\pm$0.006 & 0.45$\pm$0.008  & 0.864$\pm$0.005 &  0.68$\pm$0.009 \\
        MCTS: Majority Voting & 0.883$\pm$0.002  & 0.608$\pm$0.004 & 0.905$\pm$0.001  & 0.786$\pm$0.003 \\
        MCTS: Maximum of mean Reward & 0.776$\pm$0.0 & 0.469$\pm$0.003 & 0.854$\pm$0.003 & 0.685$\pm$0.0\\
        MCTS: Maximum of max Rewards & 0.773$\pm$0.003 & 0.485$\pm$0.003 &0.86$\pm$0.004 & 0.699$\pm$0.008 \\
        MCTS: Weighted Majority voting of mean Reward & 0.422$\pm$0.0  & 0.542$\pm$0.001 & 0.903$\pm$0.0 & 0.785$\pm$0.0\\
        MCTS: Weighted Majority voting of max Reward & 0.598$\pm$0.001  & 0.554$\pm$0.0 & 0.901$\pm$0.0 & 0.784$\pm$0.001\\
        \bottomrule 
    \end{tabular}
        \caption{Experiment results on math reasoning dataset GSM8K. The first two rows of the table correspond to the baselines: 1) no search, 2) original MCTSr implementation with ground-truth feedback. The later six rows correspond to various proposed modifications to test the effectiveness of self-feedback in search. Since each node in the MCTS tree can contain multiple reward, to get the final reward of a node we experimented with mean and max aggregation strategies. In the above table, these variations are named with suffixes "mean reward`` and "max reward`` respectively. Across all the methods without access to ground-truth feedback, the most promising alternative seems to be majority voting rather than picking the answer with maximum reward given from self-feedback} \label{tab:mctsr_appendix} 
\end{table*}


\section{ToolTalk Experiments:} \label{sec:appendix_tooltalk}
In Table~\ref{mcts_appendix}, we report results on ToolTalk with MCTS search with no search and various search strategies. Similarly, in Table~\ref{tab:dfs_appendix} we report results with DFS search inplace of MCTS.
\begingroup
\renewcommand{\arraystretch}{1.3} 
\begin{table*}[hbt!]
    \centering
    \begin{tabular}{p{2.8cm}p{2.9cm}cccc}
        \toprule
       Aggregation & Method & P & R & F1 &  Bad Action  \\
        \midrule
        \multicolumn{6}{c}{Sonnet3} \\
        \midrule
         & No Search  & 0.656$\pm$0.01  &  0.765$\pm$0.008 & 0.706$\pm$0.009 &0.275$\pm$0.013  \\
        \midrule
        \multirow{4}{*}{Value} & MCTS  & 0.502$\pm$0.02  &  0.63$\pm$0.01 & 0.559$\pm$0.015 &0.387$\pm$0.02 \\
        & + Guidelines  & 0.532$\pm$0.03  &  0.606$\pm$0.021 & 0.566$\pm$0.025 &0.367$\pm$0.041 \\
        & + ICL  & 0.708$\pm$0.012  &  0.671$\pm$0.007 & 0.689$\pm$0.004 &0.25$\pm$0.007 \\
        & + Module  & 0.754$\pm$0.014  &  0.544$\pm$0.005 & 0.632$\pm$0.007 &0.223$\pm$0.027  \\
        \hline
        \multirow{4}{*}{Majority Voting} & MCTS  & 0.445  &  0.66 & 0.532 & 0.415 \\
        & + Guidelines  & 0.469  &  0.605 & 0.528 &  0.394 \\
        & + ICL  & 0.652  &  0.668 & 0.66 & 0.247 \\
        & + Module  & 0.716  &  0.634 & 0.673 &  0.248 \\
        \hline
        \multirow{4}{*}{W.Majority Voting} & MCTS  & 0.503  &  0.672 & 0.575 & 0.377\\
        & + Guidelines  & 0.51  &  0.63 & 0.564 &  0.409 \\
        & + ICL  & 0.692  &  0.66 & 0.676 & 0.241 \\
        & + Module  & 0.768  &  0.542 & 0.636 &  0.211\\
        \midrule
        \multicolumn{6}{c}{Haiku3} \\
        \midrule
         & No Search  &  0.588$\pm$0.023  &  0.698$\pm$0.019 & 0.638$\pm$0.021 &0.311$\pm$0.024  \\
        \midrule
        \multirow{4}{*}{Value} & MCTS  & 0.567$\pm$0.012  &  0.648$\pm$0.017 & 0.605$\pm$0.013 &0.331$\pm$0.017   \\
        & + Guidelines  & 0.547$\pm$0.017  &  0.622$\pm$0.01 & 0.582$\pm$0.013 &0.338$\pm$0.013   \\
        & + ICL  & 0.622$\pm$0.031  &  0.608$\pm$0.022 & 0.615$\pm$0.026 &0.31$\pm$0.016 \\
        & + Module  & 0.623$\pm$0.022  &  0.709$\pm$0.02 & 0.663$\pm$0.02 &0.303$\pm$0.026\\
        \hline
        \multirow{4}{*}{Majority Voting} & MCTS  & 0.575  &  0.689 & 0.627 & 0.314 \\
        & + Guidelines  & 0.52  &  0.655 & 0.58 &  0.326   \\
        & + ICL  & 0.614  &  0.655 & 0.634 & 0.341 \\
        & + Module  & 0.59  &  0.702 & 0.641 &  0.314 \\
        \hline
        \multirow{4}{*}{W.Majority Voting} & MCTS  & 0.556  &  0.626 & 0.589 & 0.328 \\
        & + Guidelines  & 0.577  &  0.66 & 0.616 &  0.322 \\
        & + ICL  & 0.634  &  0.605 & 0.619 & 0.304 \\
        & + Module  & 0.604  &  0.71 & 0.653 &  0.315\\
        \bottomrule
    \end{tabular}
    \caption{On ToolTalk, we compare three strategies to build better feedback mechanisms, namely (1) augmenting
system prompt with guidelines on detecting hallucinations, (2) augmenting prompt with ICL examples, (3) augmenting feedback with an additional module to detect hallucinations. Once the MCTS search is finished, we experimented with three strategies to choose the final node with (1) highest value (2) majority voting and (3) weighted majority voting }\label{mcts_appendix}
\end{table*}

\endgroup

\begingroup
\renewcommand{\arraystretch}{1.3} 

\begin{table*}[t]
    \centering
    \begin{tabular}{p{2.6cm}p{3cm}cccc}
        \toprule
        Aggregation & Method & P & R & F1 &  Bad Action  \\
        \midrule
        \multicolumn{6}{c}{Sonnet3} \\
        \midrule
         & No Search  & 0.656$\pm$0.01  &  0.765$\pm$0.008 & 0.706$\pm$0.009 &0.275$\pm$0.013  \\
        \midrule
        \multirow{4}{*}{Value} & MCTS  & 0.474$\pm$0.014  &  0.607$\pm$0.02 & 0.532$\pm$0.014 &0.425$\pm$0.016 \\
        & + Guidelines  & 0.476$\pm$0.02  &  0.555$\pm$0.026 & 0.512$\pm$0.019 &0.427$\pm$0.027 \\
        & + ICL  & 0.706$\pm$0.02  &  0.645$\pm$0.01 & 0.674$\pm$0.014 &0.248$\pm$0.015 \\
        & + Module  & 0.756$\pm$0.026  &  0.52$\pm$0.023 & 0.616$\pm$0.024 &0.203$\pm$0.026 \\
        \hline
        \multirow{4}{*}{Majority Voting} & MCTS  & 0.328  &  0.622 & 0.43 & 0.458\\
        & + Guidelines  & 0.362  &  0.597 & 0.451 &  0.42 \\
        & + ICL  & 0.614  &  0.723 & 0.664 & 0.286 \\
        & + Module  & 0.702  &  0.605 & 0.65 &  0.23   \\
        \hline
        \multirow{4}{*}{W.Majority Voting} & MCTS  & 0.356  &  0.655 & 0.461 & 0.457\\
        & + Guidelines  & 0.374  &  0.676 & 0.482 &  0.42  \\
        & + ICL  & 0.645  &  0.655 & 0.65 & 0.292 \\
        & + Module  & 0.782  &  0.559 & 0.652 &  0.208 \\
        \midrule
        \multicolumn{6}{c}{Haiku3} \\
        \midrule
         & No Search  &  0.588$\pm$0.023  &  0.698$\pm$0.019 & 0.638$\pm$0.021 &0.311$\pm$0.024  \\
        \midrule
        \multirow{4}{*}{Value} & MCTS  & 0.562$\pm$0.029  &  0.619$\pm$0.027 & 0.589$\pm$0.028 &0.341$\pm$0.032 \\
        & + Guidelines  & 0.563$\pm$0.01  &  0.623$\pm$0.012 & 0.592$\pm$0.01 &0.325$\pm$0.018 \\
        & + ICL  &  0.66$\pm$0.015  &  0.591$\pm$0.01 & 0.623$\pm$0.005 &0.288$\pm$0.012 \\
        & + Module  & 0.61$\pm$0.005  &  0.667$\pm$0.012 & 0.637$\pm$0.007 &0.315$\pm$0.012 \\
        \hline
        \multirow{4}{*}{Majority Voting} & MCTS  & 0.571  &  0.71 & 0.633 & 0.31\\
        & + Guidelines  & 0.5  &  0.676 & 0.575 &  0.314 \\
        & + ICL  & 0.594  &  0.693 & 0.64 & 0.297 \\
        & + Module  & 0.58  &  0.685 & 0.628 &  0.346 \\
        \hline
        \multirow{4}{*}{W.Majority Voting} & MCTS  & 0.563  &  0.676 & 0.614 & 0.32 \\
        & + Guidelines  & 0.495  &  0.681 & 0.573 &  0.329 \\
        & + ICL  & 0.628  &  0.618 & 0.623 & 0.31 \\
        & + Module  & 0.593  &  0.685 & 0.636 &  0.308 \\
        \bottomrule
    \end{tabular}
    \caption{Again similar to Table 5, on ToolTalk, we compare various strategies to build better feedback mechanisms. However instead of using MCTS, we experimented with DFS search}\label{tab:dfs_appendix}
\end{table*}
\endgroup

\newpage
\section{ToolTalk System Prompts} \label{sec:appendix_tooltalk_systemprompts}

Here is the system prompts we used for main LLM that generates answer candidates for all our approaches.

\textcolor{blue}{\textit{``You are a helpful assistant who must always use one of the available external tools to successfully accomplish user task. You are also provided with some special tools to interact with the user, those are: (1) AskUserForInformation (2) FinishTask (3) AbortTask.''} }

Here is the feedback model's system prompts we used for all our approach except where we instruct the feedback model to catch hallucinations.

\textcolor{blue}{\textit{``You are a helpful critic who provides valuable feedback on solutions generated by an LLM agent. You will be given a weak or incorrect answer, and you should provide me with a feedback to correct this answer better. Analyze the answer strictly and critic, point out every flaw.''}}

For our approach, where we ask the feedback model to look for hallucination, we use the following

\textcolor{blue}{\textit{``You are a helpful critic who provides valuable feedback on solutions generated by an LLM agent. You will be given a weak or incorrect answer, and you should provide me with a feedback to correct this answer better. Analyze the answer strictly and critic, point out every flaw.}}

\textcolor{blue}{\textit{When evaluating the solutions, pay close attention to Hallucinations and Fabricated Information. Carefully evaluate if the LLM agent has hallucinated or fabricated any information, especially regarding tool parameters, but also any other potential hallucinations. Hallucinations should result in a significant decrease in the overall rating (closer to 1). Go over each tool call parameter and justify whether it is explicitly provided by the user or hallucinated.''} }

\newpage
\section{ToolTalk Error Analysis} \label{sec:appendix_tooltalk_errornalysis}

\textbf{Example 1: Showcasing Model's Hallucination problem }

\noindent \textsc{User: } \\ \texttt{Hi assistant, I want to make a new account and put an important meeting on the calendar for tomorrow.}\\ \\
\noindent \textsc{Ground Truth: }\\
\texttt{\textcolor{orange}{Ok, I can help with that. Can you give me a username, password, email, name, and phone number for the account?}}\\

\noindent \textsc{Assistant: }
\noindent \begin{lstlisting}[xleftmargin=0pt]
RegisterUser(
username: assistant_request
password: password123
email: assistant@example.com
)
\end{lstlisting}
\noindent \textsc{Tool Response: } 
\begin{lstlisting}[xleftmargin=0pt,basicstyle=\color{violet}]
session_token: e149636f-d9ca-0792
\end{lstlisting}
\noindent \textsc{Assistant: } 
\noindent \begin{lstlisting}[xleftmargin=0pt]
UserLogin(
username: assistant_request
password: password123
)
\end{lstlisting}
\noindent \textsc{Tool Response:} 
\begin{lstlisting}[xleftmargin=0pt,basicstyle=\color{violet}]
exception: Only one user can be logged in at a time. Current user is assistant_request.
\end{lstlisting}
\noindent \textsc{Assistant: }
\noindent \begin{lstlisting}[xleftmargin=0pt]
CreateEvent(
name: Important Meeting
description: All hands meeting to discuss quarterly results.
event_type: meeting
start_time: 2023-09-12 14:00:00
end_time: 2023-09-12 16:00:00
attendies: [ user1, user2]
session_token: e149636f-d9ca-0792
)
\end{lstlisting}
\noindent \textsc{Tool Response:} 
\begin{lstlisting}[xleftmargin=0pt,basicstyle=\color{violet}]
exception: null
\end{lstlisting}
\noindent \textsc{Assistant: }\\
\texttt{I have successfully created your account and added the important meeting to your calendar for tomorrow. Please let me know if you need anything else!}\\ \\

\noindent \textbf{Example 2: Showcasing Model's Hallucination problem}\\

\noindent \textsc{User: } \\ \texttt{Hey I have a morning flight tomorrow. Can you set an alarm for 8:00 am?}\\ \\
\noindent \textsc{Ground Truth: }\\
\texttt{\textcolor{orange}{Sure, I notice you're not logged in. Can you log in first?}}\\

\noindent \textsc{Assistant: }
\noindent \begin{lstlisting}[xleftmargin=0pt]
UserLogin(
username: Your username
password: Your password
)
\end{lstlisting}
\noindent \textsc{Tool Response: } 
\begin{lstlisting}[xleftmargin=0pt,basicstyle=\color{violet}]
exception: The username does not exist
\end{lstlisting}
\noindent \textsc{Assistant: }
\noindent \begin{lstlisting}[xleftmargin=0pt]
RegisterUser(
username: Your desired username
password: Your desired password
email: Your email address
)
\end{lstlisting}
\noindent \textsc{Tool Response: } 
\begin{lstlisting}[xleftmargin=0pt,basicstyle=\color{violet}]
exception: The email format is invalid
\end{lstlisting}
\noindent \textsc{Assistant: } \\ \texttt{Please provide a valid email address and desired username and password to register an account.}\\ \\

\noindent \textbf{Example 3: Showcasing Model's Hallucination problem (Hallucination at two different instances)}\\

\noindent \textsc{User: } \\ \texttt{Hey I think someone hacked my account. I can't log in anymore} \\

\noindent \textsc{Assistant }\\
\texttt{\textcolor{orange}{I'm sorry to hear that. What is your username?}}\\

\noindent \textsc{User: } \\ \texttt{It's mstein} \\

\noindent \textsc{Assistant: }
\noindent \begin{lstlisting}[xleftmargin=0pt]
SendVerificationCode(
username: mstein
email: mark@example.com
)
\end{lstlisting}
\noindent \textsc{Tool Response: } 
\begin{lstlisting}[xleftmargin=0pt,basicstyle=\color{violet}]
exception: The email is incorrect.
\end{lstlisting}

\noindent \textsc{Assistant: }\\ \texttt{Please provide the email address associated with your username mstein so I can send a verification code to reset your password.} \\

\noindent \textsc{Ground Truth: }\\
\texttt{\textcolor{orange}{Okay, I can start an account recovery process, what's your email address associated with the account?}}\\

\noindent \textsc{User: } \\ \texttt{It should be steinki89@fexter.com, hopefully the hacker didn't change it.} \\

\noindent \textsc{Assistant: }
\noindent \begin{lstlisting}[xleftmargin=0pt]
SendVerificationCode(
username: mstein
email: steinki89@fexter.com
)
\end{lstlisting}

\noindent \texttt{I have sent a 6 digit verification code to your backup email steinki89@fexter.com associated with the username mstein. Please check your inbox for the code. Once you receive the code, let me know, and I can guide you through resetting your password securely} \\

\noindent \textsc{User: } \\ \texttt{I got the code, it's 984520.} \\

\noindent \textsc{Ground Truth: }\\
\texttt{\textcolor{orange}{Okay, please enter a new password for your account.}}\\

\noindent \textsc{Assistant: }
\noindent \begin{lstlisting}[xleftmargin=0pt]
ResetPassword(
username: mstein
verification_code: 984520
new_password: NewSecurePass123!
)
\end{lstlisting}

\noindent \textsc{Tool Response: } 
\begin{lstlisting}[xleftmargin=0pt,basicstyle=\color{violet}]
status: success
exception: null
\end{lstlisting}

\noindent \textsc{Assistant: }
\noindent \begin{lstlisting}[xleftmargin=0pt]
UserLogin(
username: mstein
password: NewSecurePass123!
)
\end{lstlisting}

\noindent \textsc{Tool Response: } 
\begin{lstlisting}[xleftmargin=0pt,basicstyle=\color{violet}]
exception: null
\end{lstlisting}

\noindent \textsc{Assistant: }\\ \texttt{I have successfully reset your password and logged you into your account. Let me know if you need any other assistance securing your account further.} \\

\noindent \textbf{Example 4: Assistant has enough information but needs to deduct first} 

Below is an example prediction from the hallucination detection Module with Sonnet 3. 

\noindent \textsc{Metadata: } 
\noindent \begin{lstlisting}[xleftmargin=0pt,basicstyle=\color{magenta}]
location: London
timestamp: 2023-09-11 09:00:00
session_token: 98a5a87a-7714-b404
username: decture
\end{lstlisting}

\noindent \textsc{User: } \\ \texttt{Can you change the location for my birthday party event on sunday to "Steak and Shake" and send an email to all my friends on the invite saying "Hi, I have made some changes to the event details for my birthday party. I've decided to go to a restaurant instead- let me know if you can still make it.}\\ 

\noindent \textsc{Ground Truth: }\\
\noindent \begin{lstlisting}[xleftmargin=0pt,basicstyle=\color{orange}]
QueryCalendar(
session_token: 98a5a87a-7714-b404
start_time: 2023-09-17 00:00:00
end_time: 2023-09-17 23:59:59
)
\end{lstlisting}

\noindent \textsc{Tool Response: } 
\begin{lstlisting}[xleftmargin=0pt,basicstyle=\color{violet}]
event_id: c3463779-7861
name: Birthday Party
event_type: event
start_time: 2023-09-17 11:30:00
end_time: 2023-09-17 15:30:00
location: Chuck e cheeses
description: Happy 40th!
attendies = [decture, SuryaRani90, geoff]
\end{lstlisting}

\noindent \begin{lstlisting}[xleftmargin=0pt,basicstyle=\color{orange}]
ModifyEvent(
session_token: 98a5a87a-7714-b404
event_id: c3463779-7861
new_location: Steak and Shake
)
\end{lstlisting}

\noindent \textsc{Tool Response: } 
\begin{lstlisting}[xleftmargin=0pt,basicstyle=\color{violet}]
status: success
\end{lstlisting}

\noindent \begin{lstlisting}[xleftmargin=0pt,basicstyle=\color{orange}]
QueryUser(
session_token: 98a5a87a-7714-b404
username: SuryaRani90
)
\end{lstlisting}

\noindent \textsc{Tool Response: } 
\begin{lstlisting}[xleftmargin=0pt,basicstyle=\color{violet}]
username: SuryaRani90
email: suryarani.kumar@myfictionalemail.com
phone: 415-555-6789
name: Surya Rani Kumar
\end{lstlisting}

\noindent \begin{lstlisting}[xleftmargin=0pt,basicstyle=\color{orange}]
QueryUser(
session_token: 98a5a87a-7714-b404
username: geoff
)
\end{lstlisting}

\noindent \textsc{Tool Response: } 
\begin{lstlisting}[xleftmargin=0pt,basicstyle=\color{violet}]
username: geoff
email: geoff.smith@example.com
phone: 555-123-4567
name: Jeff Smith
\end{lstlisting}

\noindent \begin{lstlisting}[xleftmargin=0pt,basicstyle=\color{orange}]
SendEmail(
session_token: 98a5a87a-7714-b404
to: [suryarani.kumar@myfictionalemail.com, geoff.smith@example.com]
subject: Event update
body: Hi, I have made some changes to the event details for my birthday party. I've decided to go to a restaurant instead- let me know if you can still make it.
)
\end{lstlisting}

\noindent \textsc{Assistant: }\\ \texttt{Could you please provide the event_id of your birthday party event that needs to be modified?} \\

\noindent \textbf{Example 5: Asking for unnecessary information} 

\noindent \textsc{Metadata: } 
\noindent \begin{lstlisting}[xleftmargin=0pt,basicstyle=\color{magenta}]
location: London
timestamp: 2023-09-10 09:00:00
session_token: 98a5a87a-7714-b404
username: justinkool
\end{lstlisting}

\noindent \textsc{User: } \\ \texttt{I need to cancel my meeting with Alicia tomorrow. Can you check the details for the event?}\\

\noindent \textsc{Ground Truth: }\\
\noindent \begin{lstlisting}[xleftmargin=0pt,basicstyle=\color{orange}]
QueryCalendar(
session_token: 98a5a87a-7714-b404
session_token: 98a5a87a-7714-b404
start_time: 2023-09-11 00:00:00
end_time: 2023-09-11 23:59:59
)
\end{lstlisting}

\noindent \textsc{Tool Response: } 
\begin{lstlisting}[xleftmargin=0pt,basicstyle=\color{violet}]
event_id: 5480d946-2381
name: Justin:Alicia
event_type: meeting
start_time: 2023-09-11 10:00:00
end_time: 2023-09-11 10:30:00
location: Conference Room 1
description: Talk about anything. Move to your convenience.
\end{lstlisting}

\texttt{\textcolor{orange}{I found a meeting tomorrow at 10 am with the title Justin:Alicia. The description says "Talk about anything. Move to your convenience." The attendees are you and Alicia (aliciab)}}\\

\noindent \textsc{Assistant: }\\ \texttt{Could you please provide the date, start time, and end time of the meeting with Alicia that you need to cancel? Any other details like the meeting title or attendees would also be helpful to identify the correct event?} \\

\noindent \textbf{Example 6: Asking for User Confirmation}

The following is a prediction from the Hallucination detection module approach \\

\noindent \textsc{Metadata: } 
\noindent \begin{lstlisting}[xleftmargin=0pt,basicstyle=\color{magenta}]
location: San Francisco
timestamp: 2023-09-11 09:00:00
\end{lstlisting} 

\noindent \textsc{User: } \\ \texttt{Hey can you help me check my account information?}\\

\noindent \textsc{Assistant: }\\ 
\textcolor{orange}{\texttt{Sure, I see you're not logged in. Can you provide you username and password?}} \\

\noindent \textsc{User: } \\ \texttt{Yeah, it's justinkool and justforkicks123}\\

\noindent \textsc{Ground Truth: } 
\noindent \begin{lstlisting}[xleftmargin=0pt,basicstyle=\color{orange}]
UserLogin(
username: justinkool
password: justforkicks123
)
\end{lstlisting}

\noindent \textsc{Tool Response: } 
\begin{lstlisting}[xleftmargin=0pt,basicstyle=\color{violet}]
event_id: 5480d946-2381
session_token: e149636f-d9ca-0792
\end{lstlisting}
\noindent \begin{lstlisting}[xleftmargin=0pt,basicstyle=\color{orange}]
GetAccountInformation(
session_token: e149636f-d9ca-0792)
\end{lstlisting}

\noindent \textsc{Tool Response: } 
\begin{lstlisting}[xleftmargin=0pt,basicstyle=\color{violet}]
username: justinkool
email: justintime@fmail.com
phone: 123-456-7890
name: Justin Kool
\end{lstlisting} 

\noindent \textsc{Assistant: }
\noindent \begin{lstlisting}[xleftmargin=0pt]
UserLogin(
username: justinkool
password: justforkicks123
)
\end{lstlisting}
\noindent \textsc{Tool Response: } 
\begin{lstlisting}[xleftmargin=0pt,basicstyle=\color{violet}]
event_id: 5480d946-2381
session_token: e149636f-d9ca-0792
\end{lstlisting}

\noindent \texttt{For security reasons, before I can check your account information, I need you to provide some additional verification details that only the real account owner would know, such as: 1) The email address associated with your account 2) Your phone number in the xxx-xxx-xxxx format Please provide this additional verification information, and I'll be happy to check your account details once verified. Let me know if you have any other questions!"
} \\

\noindent \textbf{Example 7: Asking for unnecessary information}

\noindent \textsc{Metadata: } 
\noindent \begin{lstlisting}[xleftmargin=0pt,basicstyle=\color{magenta}]
location: San Francisco
timestamp: 2023-09-11 09:00:00
session_token: 98a5a87a-7714-b404
username: decture
\end{lstlisting} 

\noindent \textsc{User: } \\ \texttt{Hey I need to remember to buy a birthday gift for my spouse before I go home. Can you set an alarm for 2:30 to remind me?}\\

\noindent \textsc{Assistant: }
\noindent \begin{lstlisting}[xleftmargin=0pt,basicstyle=\color{orange}]
AddAlarm(
session_token: 98a5a87a-7714-b404
time: 14:30:00
)
\end{lstlisting}

\noindent \textsc{Tool Response: } 
\begin{lstlisting}[xleftmargin=0pt,basicstyle=\color{violet}]
alarm_id: 5bff-dd80
\end{lstlisting}

\textcolor{orange}{\texttt{Sure, I've set an alarm for 2:30 PM to remind you to buy a birthday gift for your spouse.}} \\

\noindent \textsc{User: } \\ \texttt{Thanks! Can you also check my calendar to see if I have any other events today?}\\

\noindent \textsc{Assistant: }
\noindent \begin{lstlisting}[xleftmargin=0pt,basicstyle=\color{orange}]
QueryCalendar(
session_token: 98a5a87a-7714-b404
start_time: 2023-09-11 00:00:00
end_time: 2023-09-11 23:59:59
)
\end{lstlisting}

\noindent \textsc{Tool Response: } 
\begin{lstlisting}[xleftmargin=0pt,basicstyle=\color{violet}]
event_id: 29496535-b409
name: Dentist Appointment
event_type: event
start_time: 2023-09-11 15:00:00
end_time: 2023-09-11 17:00:00
location: dental office
description: null
attendees: null
\end{lstlisting} 

\textcolor{orange}{\texttt{Sure, you have a dentist appointment at 4:00 PM today.}} \\

\noindent \textsc{Assistant: }
\noindent \texttt{What time range would you like me to check your calendar for events today? Please provide the start and end times in the format \%Y-\%m-\%d \%H:\%M:\%S."
} \\

\end{document}